\documentclass{article} 

\usepackage{gen2_iclr2026,times}

\usepackage{amsmath,amsfonts,bm}

\def\eqref#1{equation~\ref{#1}}

\def\1{\bm{1}}

\DeclareMathAlphabet{\mathsfit}{\encodingdefault}{\sfdefault}{m}{sl}
\SetMathAlphabet{\mathsfit}{bold}{\encodingdefault}{\sfdefault}{bx}{n}

\usepackage{hyperref}
\usepackage{url}
\usepackage{subfigure}
\usepackage{dsfont}
\usepackage{graphicx}
\usepackage{amsmath}

\usepackage{amssymb}
\usepackage{bm}
\usepackage{float}
\usepackage{booktabs}
\usepackage{multirow}
\usepackage{makecell} 
\usepackage{verbatim}

\title{Contact-Guided 3D Genome Structure Generation of \textit{E. coli} via Diffusion Transformers\\[-0.5em]}

\author{
{\normalfont
\begin{tabular*}{\linewidth}{@{\extracolsep{\fill}}ll}
\textbf{Mingxin Zhang}\thanks{Equal contribution.} \thanks{Corresponding author.} &
\textbf{Xiaofeng Dai}\footnotemark[1] \\
The University of Tokyo & University of Michigan \\
\texttt{m.zhang@hapis.k.u-tokyo.ac.jp} &
\texttt{xiaofend@umich.edu} \\
\\
\textbf{Yu Yao}\footnotemark[1] &
\textbf{Ziqi Yin} \\
The University of Tokyo & Waseda University \\
\texttt{yao@ms.k.u-tokyo.ac.jp} &
\texttt{yinziqi2001@toki.waseda.jp}
\end{tabular*}
}
}

\iclrfinalcopy 
\begin{document}

\maketitle

\begin{abstract}
In this study, we present a conditional diffusion–transformer framework for generating ensembles of three-dimensional \textit{Escherichia coli} genome conformations guided by Hi-C contact maps. Instead of producing a single deterministic structure, we formulate genome reconstruction as a conditional generative modeling problem that samples heterogeneous conformations whose ensemble-averaged contacts are consistent with the input Hi-C data. A synthetic dataset is constructed using coarse-grained molecular dynamics simulations to generate chromatin ensembles and corresponding Hi-C maps under circular topology. Our models operate in a latent diffusion setting with a variational autoencoder that preserves per-bin alignment and supports replication-aware representations. Hi-C information is injected through a transformer-based encoder and cross-attention, enforcing a physically interpretable one-way constraint from Hi-C to structure. The model is trained using a flow-matching objective for stable optimization. On held-out ensembles, generated structures reproduce the input Hi-C distance-decay and structural correlation metrics while maintaining substantial conformational diversity, demonstrating the effectiveness of diffusion-based generative modeling for ensemble-level 3D genome reconstruction.
\end{abstract}

\section{INTRODUCTION}
Since the discovery of the DNA double helix by Watson and Crick \cite{watson1953molecular}, genomic information has been understood primarily as a linear nucleotide sequence. Yet in living cells, DNA is organized as a three-dimensional (3D) polymer with hierarchical structure. A growing body of evidence indicates that 3D genome organization is closely linked to fundamental cellular processes, including replication, transcription, and DNA repair \cite{misteli2020self}. Consequently, reconstructing genome structure is important both for basic biology and for applications in biotechnology and medicine.

Direct visualization of chromosome conformations at high resolution remains challenging for most systems due to the limitation of imaging techniques \cite{lakadamyali2020visualizing}. Chromosome conformation capture assays—most prominently Hi-C—provide an alternative by measuring population-averaged contact frequencies between genomic fragments via high-throughput sequencing \cite{lieberman2009comprehensive}. However, a Hi-C contact map is an indirect measurement: it reports interaction frequencies rather than a unique 3D structure, and the same contact map can in principle arise from multiple distinct conformational ensembles.

A wide range of computational approaches have been developed to infer 3D genome structure from Hi-C data, including optimization-based reconstruction, multidimensional scaling, polymer simulations, and machine learning models \cite{meluzzi2020computational, lee2025unraveling}. Despite notable progress, most existing methods are effectively deterministic: they produce a single consensus conformation that best matches the observed contacts. This paradigm overlooks the intrinsic heterogeneity of chromosome organization, where a distribution of conformations—rather than a single structure—gives rise to the observed population Hi-C signal. Ensemble-based methods have been proposed, but they are often computationally expensive and can be difficult to scale \cite{li2024hi}.

We instead view genome reconstruction as a conditional generative modeling problem: given a Hi-C map, the goal is to sample a distribution of physically plausible 3D conformations whose aggregated contacts are consistent with the measurement. Such a formulation naturally represents uncertainty and heterogeneity, and offers a path toward amortized inference—once trained, the model can generate ensembles efficiently for new inputs.

In this work, we develop a transformer-based diffusion model to generate ensembles of \textit{Escherichia coli} (\textit{E. coli}) 3D genome conformations conditioned on published Hi-C data. We build on Diffusion Transformers (DiTs) \cite{Peebles2022DiT}, which are well matched to chromatin representations as sequences of 3D coordinates and can capture long-range dependencies more flexibly than convolutional U-Net backbones used in prior work \cite{schuette2025chromogen}. We focus on bacteria for two reasons. First, bacterial chromosomes offer comparatively well-studied physical and biological constraints, enabling more stringent plausibility checks on generated structures. Second, prokaryotes span diverse genome organization regimes, providing a natural testbed for evaluating transferability across species. Together, these properties make bacterial systems a meaningful setting for developing scalable generative models for 3D genome generation in the future.

\section{RELATED WORKS}
Prior work most closely related to ours studies the reconstruction of 3D genome architecture from Hi-C contact maps, where the goal is to infer physically plausible coordinates from indirect and noisy interaction measurements. FLAMINGO is representative of optimization-based reconstruction methods that impose explicit geometric structure: by exploiting properties of Euclidean distance geometry (e.g., low-rank structure in distance representations) and using scalable solvers, it recovers 3D coordinates from contact-derived constraints and scales to high-resolution settings via hierarchical reconstruction \cite{wang2022reconstruct}. In contrast, CHROMFORMER applies a transformer to map Hi-C contact matrices to 3D coordinates \cite{valeyre2022chromformer}. The attention mechanism is well matched to Hi-C because it can model long-range, non-local dependencies across genomic loci and flexibly integrate multi-scale interaction patterns, which are central to chromatin folding. Notably, both FLAMINGO and CHROMFORMER are primarily formulated to output a single consensus conformation for a given input contact map, whereas our work focuses on generating an ensemble of plausible 3D structures to better reflect the intrinsic structural variability consistent with the same Hi-C observations. 
Generative modeling conditioned on sequence-derived inputs has also been applied to chromatin organization. ChromoGen tackles the problem of generating chromatin organization conditioned on DNA sequence and chromatin accessibility, which also addresses the cell-to-cell heterogeneity with generative model \cite{schuette2025chromogen}. However, our work is centered on Hi-C–driven 3D generation, where predicting explicit 3D conformations is preferable because it yields a concrete geometric representation that can be directly interrogated for spatial relationships (e.g., looping, domain packing, and locus–locus proximity) and supports downstream structure-based analyses beyond what is available from an genome organization alone.



\section{METHODS}
\subsection{Chromosome Structure Data Simulation}
Because direct ground-truth 3D conformations of the \textit{E. coli} chromosome are extremely scarce—and when available, are typically obtained under experimental conditions that do not align with the wide variety of published Hi-C datasets—we rely on physics-based simulations to generate matched structure ensembles for training and evaluation. To obtain training data that respect polymer physics while remaining anchored to experimental measurements, we generated synthetic chromosome conformations using coarse-grained molecular dynamics (MD) simulations that correspond to published \textit{E. coli} Hi-C maps. We simulated the chromosome as a confined polymer with basic physical constraints—chain connectivity, excluded volume, and circular topology—inside a rectangular box chosen to approximate the dimension of an \textit{E. coli} cell (900 nm × 900 nm × 2000 nm). The experimental Hi-C maps are processed at 5k base paries (bps) resolution, yielding 928 × 928 contact matrices (4.6M bps \textit{E. coli} genome). To match this resolution, we constructed a coarse-grained polymer with 10 beads corresponding to one 5 kb Hi-C bin in the MD model.

Experimental information is incorporated during initialization: Hi-C–derived interactions are introduced as restraints to bias the polymer toward biologically reasonable conformations \cite{wasim2021hi}, together with additional setup of gene expression patterns and ribosome localization. After initialization, these restraints are gradually released and we run unbiased MD for $2 \times10^6$ integration steps, sampling conformations every 1,000 steps for data augmentation. To reflect ongoing replication in growing cells, we allow between one and two chromosome copies via a replication factor G, setting the total bead count to $928 \times 10 \times G$ (two chromosome arms and replicated copies). For data augmentation, the final structure is down-sampled by factor of 5 from the simulated structure resulting in a resolution of 2.5k bps per bead (each Hi-C bin corresponds to 2 beads).

\subsection{ResNet VAE}
\begin{figure}[H]
  \centering
  \includegraphics[width=0.6\linewidth]{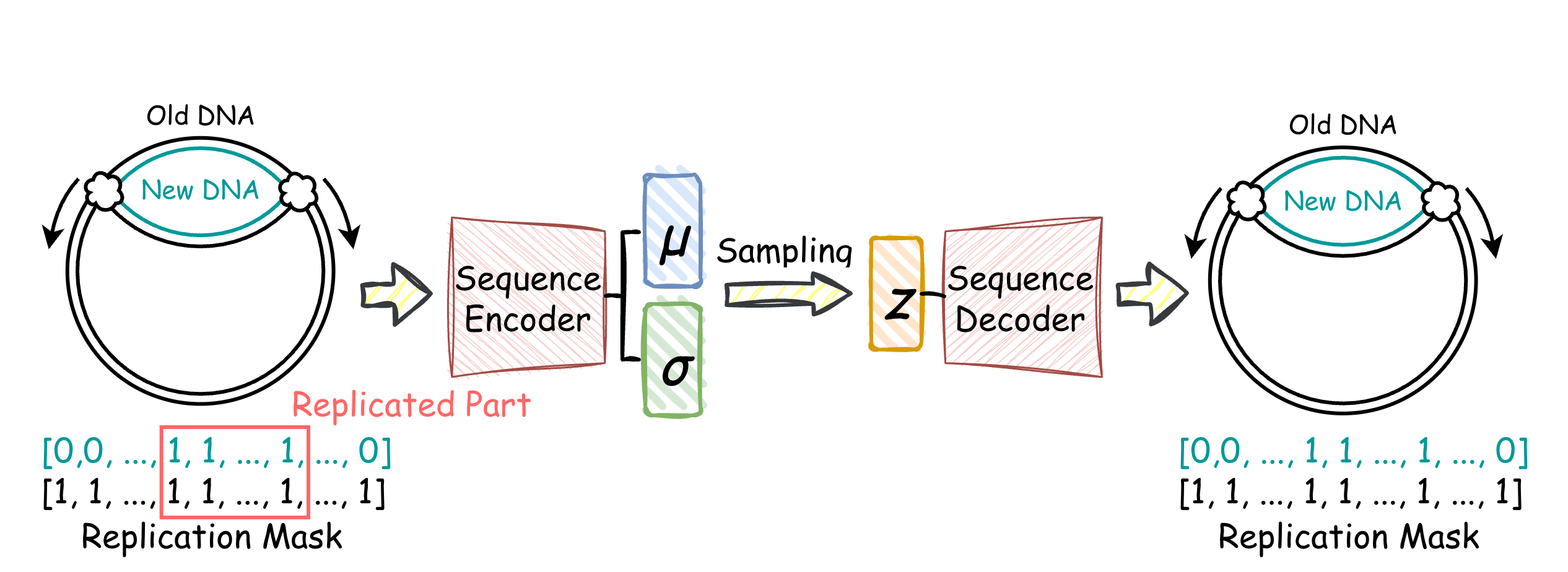}
  \caption{VAE to obtain the latent representation of 3D chromosome structures. Because chromatin may be undergoing replication and thus may not be fully replicated, a mask is used to denote bead presence. The mask equals 1 for all positions on the parental chain, and equals 1 (replicated) or 0 (unreplicated) on the new chain.}
  \label{fig:vae}
\end{figure}

To achieve efficient and stable generation, we constructed a conditional diffusion model within the latent diffusion framework \cite{rombach2022high}, which significantly reduces computational cost while improving training stability by operating in a compressed representation space. To obtain latent representations that are compatible with this model architecture, we further require a VAE to encode three-dimensional structures into a latent space suitable for diffusion-based generation. Because the genomic topology of \textit{E. coli} is relatively simple and forms a fixed circular structure, and because the beads have already been ordered according to the Hi-C index, we adopted a straightforward 1D ResNet18 VAE (shown in Fig. \ref{fig:vae}) to encode the 3D structural sequence into the latent space. To preserve the alignment between beads and the corresponding rows and columns of the Hi-C matrix, the sequence length is kept unchanged; the VAE is thus used to learn latent representations rather than to perform length compression.

Because chromatin may be undergoing replication, certain segments can exhibit branching structures. To guide the model in generating chromatin conformations during replication, we introduced two replication masks that indicate the presence or absence of beads at each position on the newly synthesized and parental chains, respectively. Accordingly, the VAE is trained not only with a coordinate reconstruction loss $\mathcal{L}_{\mathrm{coord}}$ and a KL-divergence loss $\mathcal{L}_{\mathrm{KL}}$, but also with an additional mask reconstruction loss $\mathcal{L}_{\mathrm{mask}}$:
    $\mathcal{L}=\mathcal{L}_{\mathrm{coord}}+
    \lambda_{\mathrm{mask}}\,\mathcal{L}_{\mathrm{mask}}+
    \lambda_{\mathrm{KL}}\,\mathcal{L}_{\mathrm{KL}}$, where $\mathcal{L}_{\mathrm{coord}}$ is calculated by

    $\mathcal{L}_{\mathrm{coord}}=\mathbb{E}_{x}\left[
        \mathbb{E}_{i \sim m}\left[
            \left(\hat{x}^{\mathrm{coord}}_i-x^{\mathrm{coord}}_i\right)^2
        \right]
    \right]$
and $\mathcal{L}_{\mathrm{mask}}$ is binary cross entropy loss between the predicted replication mask and the real replication mask. This means that the mean squared error is computed only at positions where the mask is active and then averaged over the valid positions. As a result, the loss focuses on physically meaningful beads while remaining invariant to the number of valid beads in each sample, thereby avoiding potential bias arising from differences in bead counts across samples of different replication stages. With this VAE, the latent vectors generated by the diffusion model can be decoded back into the corresponding 3D structures.

\subsection{CrossDiT-based DiffBacChrom}

\begin{figure}
  \includegraphics[width=\linewidth]{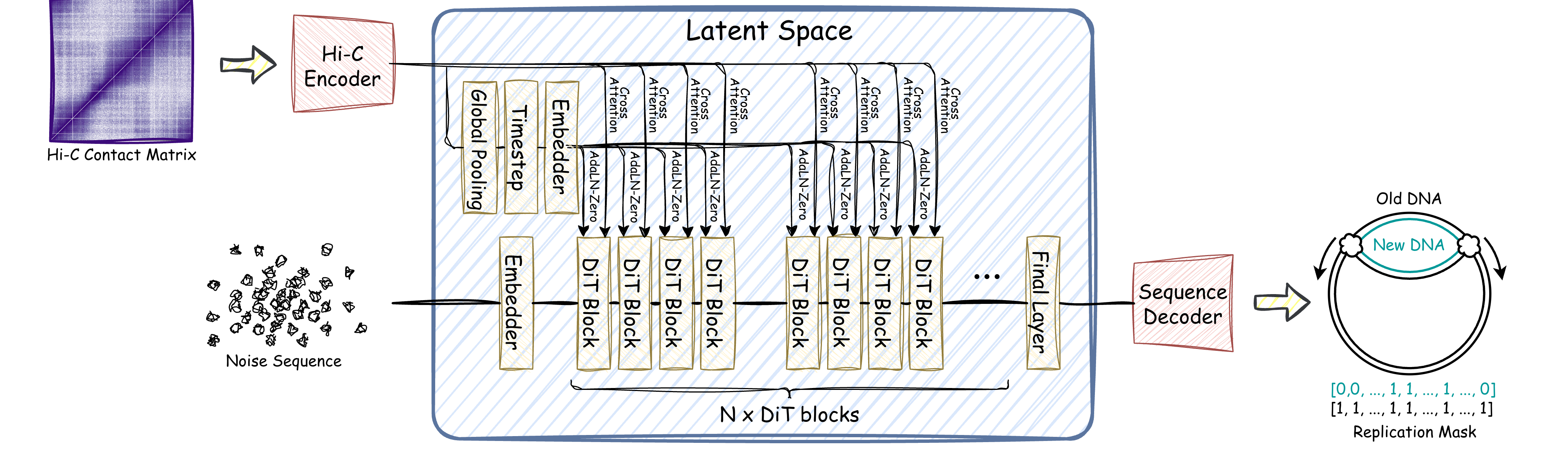}
  \caption{CrossDiT-based DiffBacChrom. The diffusion model works in the latent space, and the pre-trained VAE reconstructs the chromosome structure sequence. A Hi-C encoder transforms Hi-C maps into conditional tokens and injects them into the diffusion model.}
  \label{fig:dit}
\end{figure}

From the perspective of explicit physical constraints, our goal is to generate 3D structures that satisfy the conditions imposed by the Hi-C matrix. In this sense, the Hi-C data act as an external field, a form of constraint that determines the structure of the structure sequence without being influenced by the sequence in return. Naturally, this leads us to consider the CrossDiT architecture \cite{chen2023pixartalpha}, in which conditional injection via cross-attention ensures that the influence of the condition is strictly unidirectional. The CrossDiT model used in this study is shown in Fig. \ref{fig:dit}. A transformer-based Hi-C encoder converts the 2D Hi-C matrix into conditional embeddings $z_c$ that guide the generation process. Specifically, the Hi-C matrix is treated as a sequence along the row dimension, with columns serving as the feature dimension. This design preserves the original sequence length, ensuring bin-wise embeddings, where each element encodes the interaction between a given row and column in the Hi-C matrix. The global average pooling of $z_c$ is combined with the timestep embedding $t$ like $c=t+\widetilde{z}_c$. The global conditional embedding $c$ is injected into each DiT block via AdaLN-Zero and self-attention, following the design of the basic DiT architecture \cite{Peebles2022DiT} to further stabilize the training process.

A latent diffusion framework is adopted, in which the model operates on latent tokens $x$ rather than raw structures. The diffusion process generates the latent representation of the structure starting from a noise sequence. To ensure compatibility between the generative and conditional pathways, the sequence length of $x$ is set to match that of the conditional embedding $z_c$. Then AdaLN modulated $x$ serves as $Q$ and $z_c$ serves as $K$ and $V$ of cross-attention in each DiT block.

In this formulation, $x$ dominates the query $Q$, so that the model retrieves information from the condition, while the conditional representations themselves are not updated. This asymmetry aligns with the physical interpretability of the constraint. In addition, the lengths of $x$ and $z_c$ are consistent with the sequence length and the dimensions of the Hi-C matrix, and this structural consistency ensures bin-level alignment.

To improve training efficiency and stability, we follow the advanced DiT design and adopt a flow-matching framework \cite{esser2024scalingrectifiedflowtransformers, liu2022flow} rather than the original DDPM formulation \cite{ho2020denoising}, enabling more direct and stable optimization of the generative dynamics.

\section{EXPERIMENTS}

\subsection{Data Preparation}
\textbf{Ensemble pairing and synthetic Hi-C construction.} For the training set, we generated a pool of conformations by MD simulation on GROMACS \cite{Abraham2015GROMACS} and then formed structure–Hi-C pairs by aggregating contacts across a sampled ensemble. Concretely, we randomly selected N = 500 structures to define one ensemble; bin–bin contacts were aggregated across this ensemble to compute a single Hi-C contact matrix representing that condition. The training set contains 65 such ensembles, producing a total of 65 $\times$ 500 = 32500 structures. Each training sample is formed as a pair consisting of an individual structure and the Hi-C map corresponding to its ensemble, yielding multiple plausible 3D conformations associated with the same condition-specific Hi-C map. (Because N is far smaller than the number of cells contributing to an experimental Hi-C map, the grouping choice substantially affects the resulting synthetic contact patterns; we treat this as a key design choice in the pipeline.) We also constructed a test set comprising 10 ensembles with a total of 5000 samples to evaluate the performance of models.

\textbf{Unified representation with replication masks.} Each conformation is stored in a consistent tabular form that interfaces cleanly with the learning pipeline. Each row corresponding to a Hi-C index: 
\[  [x_1,y_1,z_1, m_1,\; x_2,y_2,z_2, m_2],    \]
where $(x,y,z)$ are coordinates, subscripts 1 and 2 denote the parental and replicated chromosomes, and $m$ is a binary replication mask indicating whether a bead is replicated. Because each Hi-C index corresponds to two beads, each template chain contains a total of 928 $\times$ 2 beads. To keep alignment with the Hi-C matrix, the two beads associated with the same Hi-C index are jointly treated as input features. Therefore, the model input has a dimensionality of 928 $\times$ 16, with per-token channels encoding the bead coordinates and replication masks for both chromosomes.

\textbf{Preprocessing.} To reduce nuisance variability and improve generalization, we applied position normalization, scale normalization, and random rotation to each sample. Specifically, we first translated each structure so that its center of mass is located at the origin, thereby eliminating translational degrees of freedom.

Secondly, we normalized each individual structure by rescaling the bead coordinates so that the mean Euclidean norm of all bead positions in the structure is equal to 1. This scale normalization removes variations in absolute size across structures, allowing the model to focus on learning relative geometric configurations rather than being biased by global scale differences.

Finally, we applied a random rotation to each sample to enforce rotational invariance and prevent the model from memorizing absolute orientations, thereby improving generalization. Each rotation was generated by sampling a unit quaternion uniformly from the 3D rotation group SO(3), which provides an efficient and unbiased parameterization of 3D rotations. The quaternion was then converted into a 3 $\times$ 3 rotation matrix and applied to all bead coordinates of the structure. An independent random rotation was used for each sample, ensuring that the model learns rotationally invariant structural features rather than relying on fixed global orientations. Random rotation was applied only to training data. 

Each training sample consists of a pair formed by an individual structure and the Hi-C map corresponding to the ensemble it belongs to.

\subsection{Model Training}
All models were trained on a single NVIDIA GH200 GPU. The hidden size of the VAE was set to 128, resulting in a 3.6M VAE encoder. The learning rate was set to $1 \times 10^{-4}$ with a batch size of 32, and the model was trained for 50 epochs. Because the KL-divergence term is only required to encourage a more regularized and smooth latent space, while ensuring that the model primarily relies on the reconstruction loss to preserve structural information, the KL-divergence term $\mathcal{L}_{\mathrm{KL}}$ was weighted relatively small by $5 \times 10^{-3}$, while the mask reconstruction loss $\mathcal{L}_{\mathrm{mask}}$ was assigned a weight of 1.0.

The learning rate of CrossDiT was set to $1 \times 10^{-4}$ with a batch size of 8, and the model was trained for 50 epochs. Because diffusion models operating in latent space require the latent variables to be properly normalized to a consistent scale so that the noise schedule and denoising dynamics are well calibrated, we computed a latent scale after training the VAE. We traversed all samples in the training set to estimate the distribution of latent representations and obtained a latent scale of 1.335256, which was then used for both training and inference of the DiT model.

\begin{table}[ht]
\caption{Backbone model configurations with different sizes.}
\label{tab:config}
\begin{center}
\begin{tabular}{rrrrr}
\bf Model  &\bf Depth &\bf hidden size  &\bf num heads &\bf Params\\
\midrule
CrossDiT-S  &12 &384   &6 &45M\\ 
CrossDiT-L  &24 &1024    &16  &634M\\
\end{tabular}
\end{center}
\end{table}

To investigate the impact of model capacity on generation quality, we followed the standard DiT configuration and train two CrossDiT variants with different scales (CrossDiT-S/L), as summarized in the Table \ref{tab:config}. Correspondingly, the Hi-C encoder used for conditional injection was configured with a depth of 8, while its hidden size and number of attention heads were matched to those of the corresponding backbone network.

\subsection{Generation}

In addition to validation on the test set, we further evaluated whether the model can generate diverse samples that collectively satisfy the input Hi-C constraints at the ensemble level. For each ensemble in the test set, we used the corresponding Hi-C matrix as the condition to generate 500 structures as a new ensemble, matching the number of structures per ensemble in the test set. We then computed a new Hi-C matrix from the generated ensemble and compared it with the original input Hi-C matrix. This procedure allows us to assess whether the model can learn ensemble-level constraints from paired one-to-one training data and reproduce the relationship in generation.

During generation, we employed classifier-free guidance (CFG) \cite{ho2022classifier}, a technique that adjusts the strength of conditional guidance to control the trade-off between condition fidelity and sample diversity. In our task, while we aim for the generated structures to satisfy the Hi-C constraints at the ensemble level, we also want individual samples to remain as diverse as possible. Therefore, the CFG scale was set to 1.0, which means that conditional and unconditional predictions are combined without amplifying the conditional signal. Unlike without CFG, the conditional pathway is still explicitly involved in generation, but without enforcing stronger guidance that could reduce diversity. For rectified flow sampling, the number of sampling steps was set to 50.

\section{RESULTS}

\begin{figure}
    \centering
    \subfigure[]{
        \begin{minipage}[b][5.4cm][c]{0.4\linewidth}
            \centering
            \includegraphics[width=\linewidth]{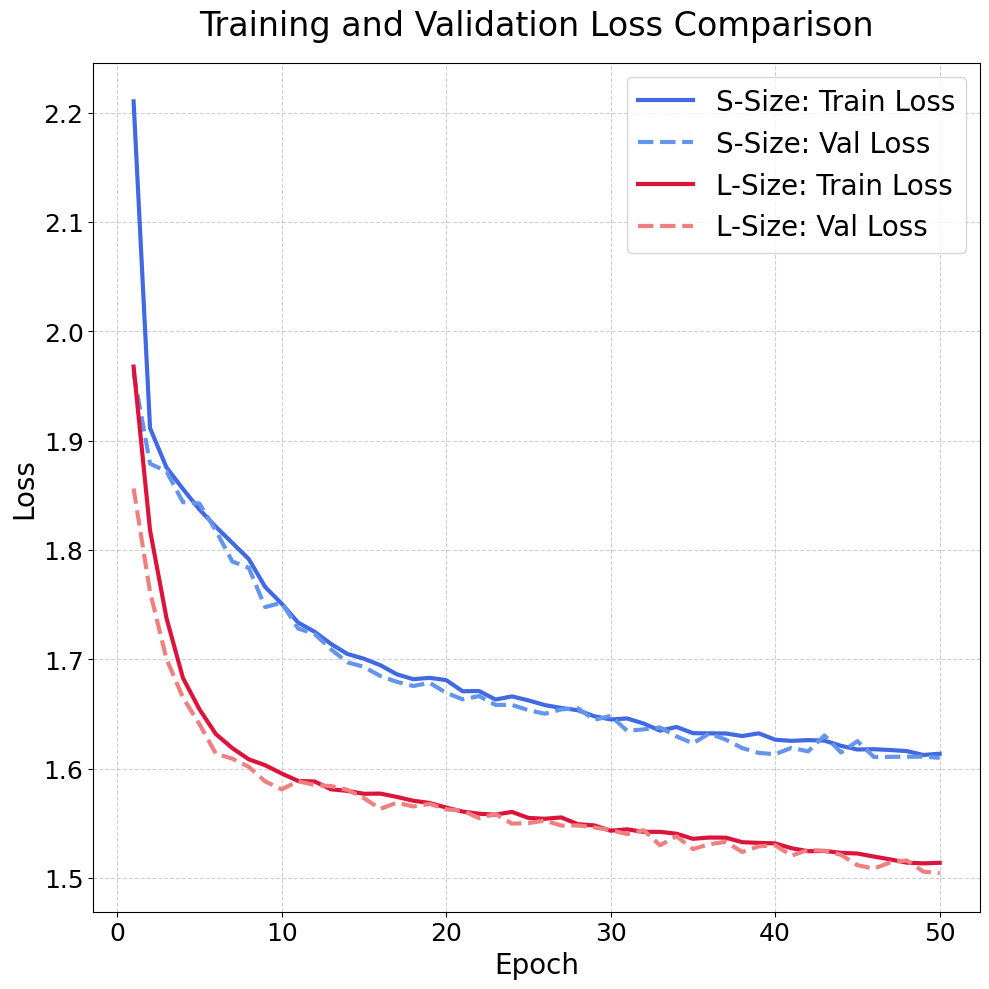}
        \end{minipage}
        \label{fig:loss}
    }
    \subfigure[]{
        \begin{minipage}[b][4.5cm][c]{0.27\linewidth}
            \centering
            \includegraphics[width=\linewidth]{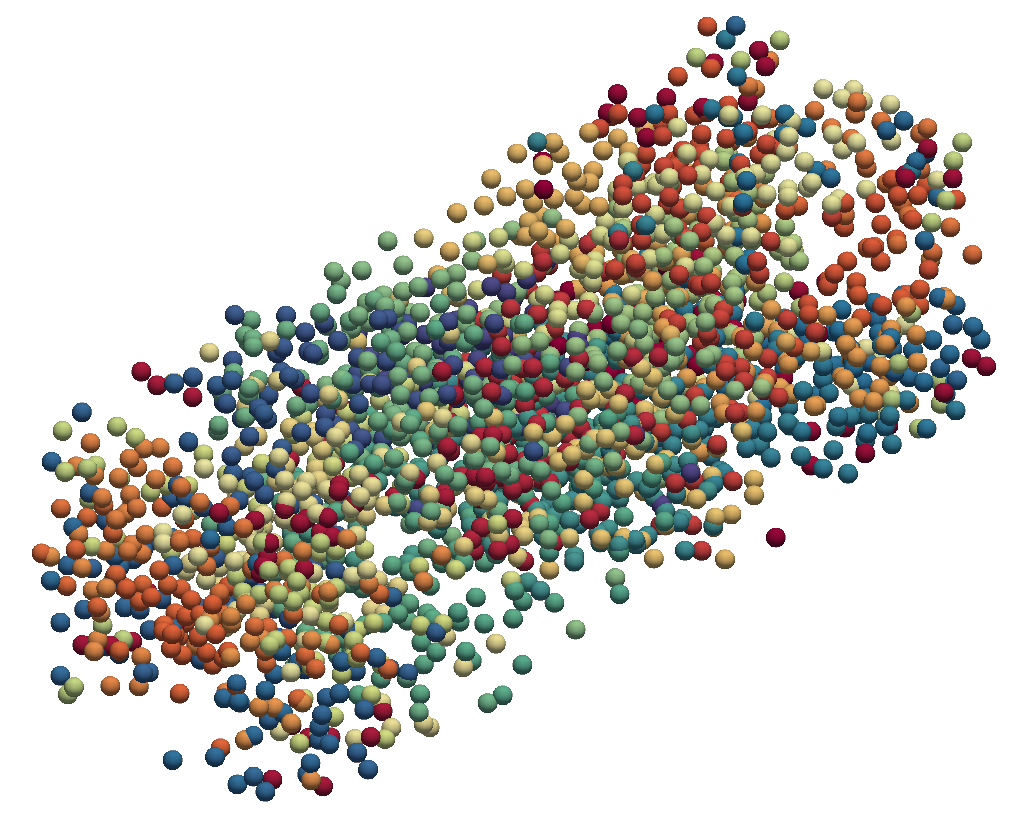}
        \end{minipage}
        \label{fig:gt_vis}
    }
    \subfigure[]{
        \begin{minipage}[b][4.5cm][c]{0.27\linewidth}
            \centering
            \includegraphics[width=\linewidth]{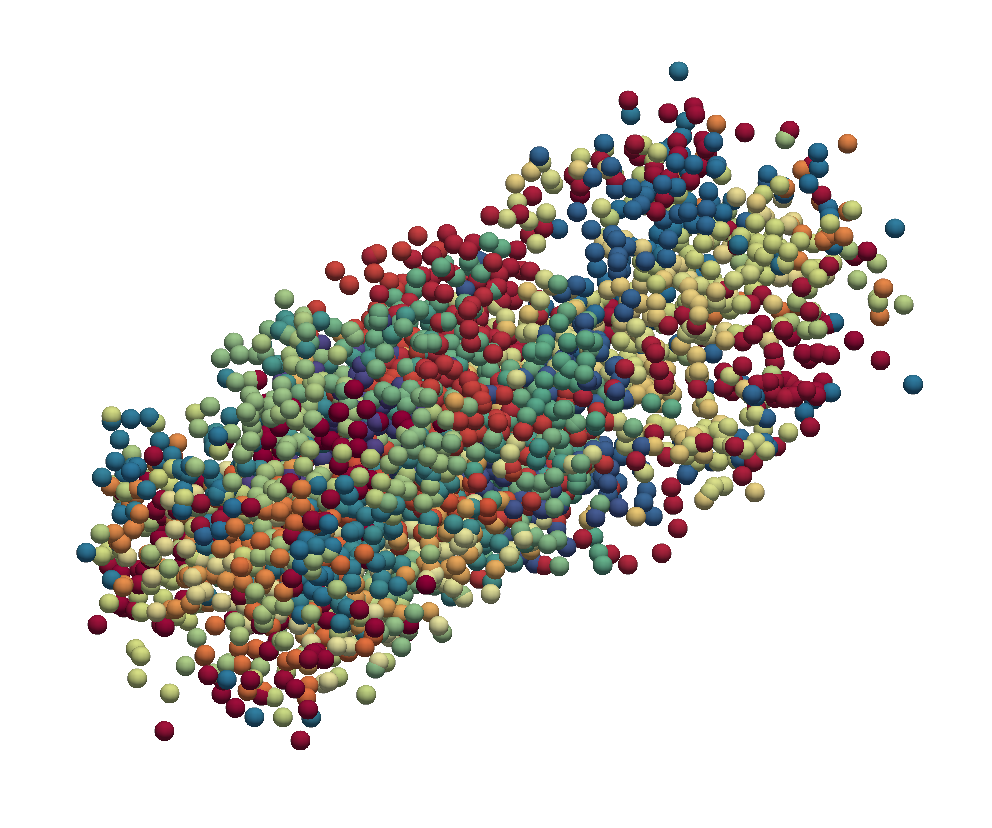}
        \end{minipage}
        \label{fig:gen_vis}
    }
    
  \caption{(a) Training loss and validation loss of CrossDiT-S/L; (b) Visualization of example chromosome structure from MD simulation; (c) Example generated structure guided by the Hi-C matrix corresponding to (b).}
  \label{fig:vis}
\end{figure}

For each generated ensemble, we compute the corresponding Hi-C map using the same procedure applied to the dataset samples. Because the structures fed into the model are scale-normalized, we first recover the average coordinate norm in the original sample space and rescale the contact distance threshold accordingly. This ensures that the Hi-C maps computed from generated structures are directly comparable to the original Hi-C maps and are not affected by the normalization scale.

\begin{figure}
    \centering
    \subfigure[]{
        \begin{minipage}[b]{0.4\linewidth}
            \includegraphics[width=\linewidth]{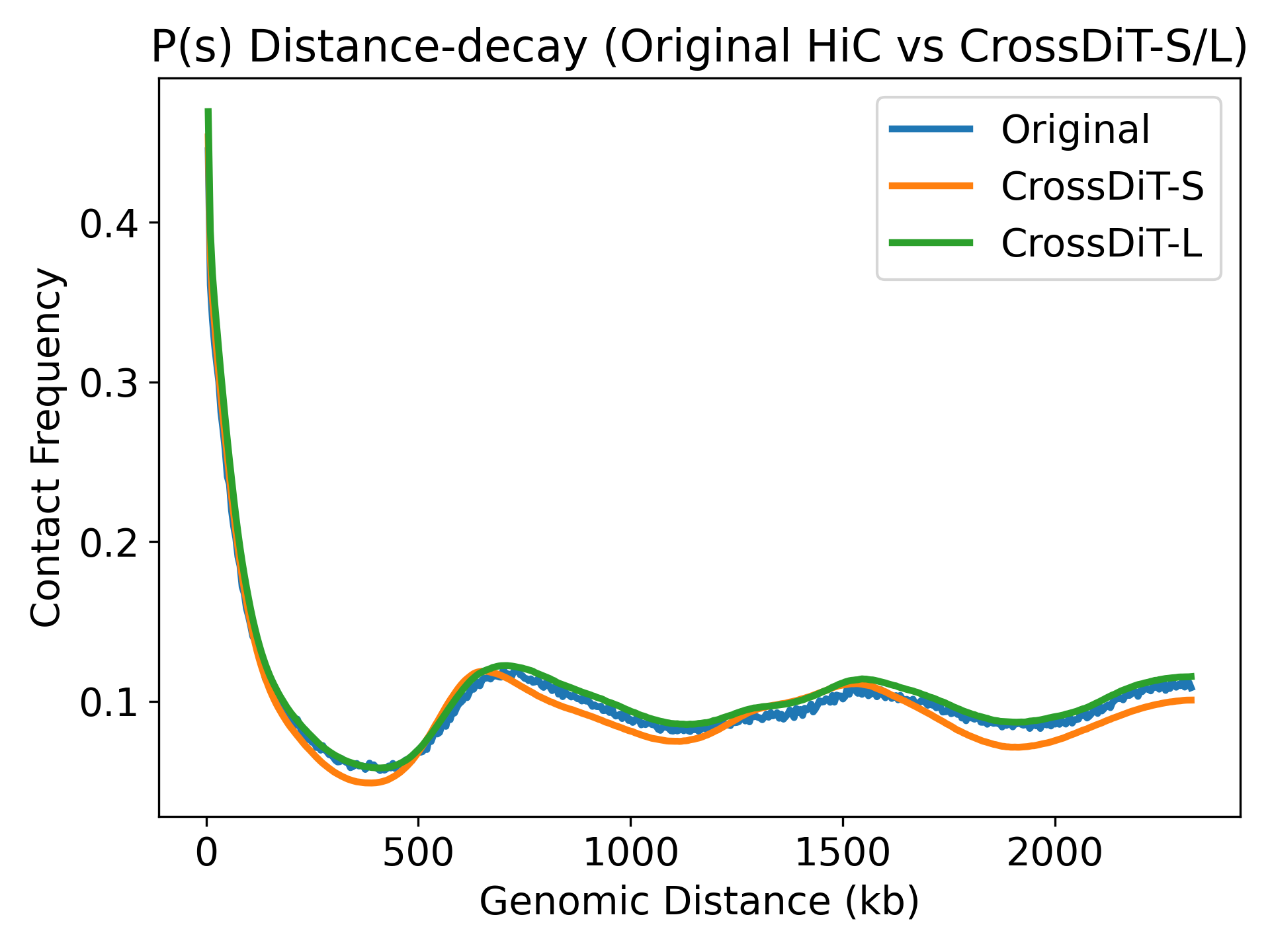} 
        \end{minipage}
        \label{fig:ps}
    }
    \subfigure[]{
        \begin{minipage}[b]{0.5\linewidth}
            \includegraphics[width=\linewidth]{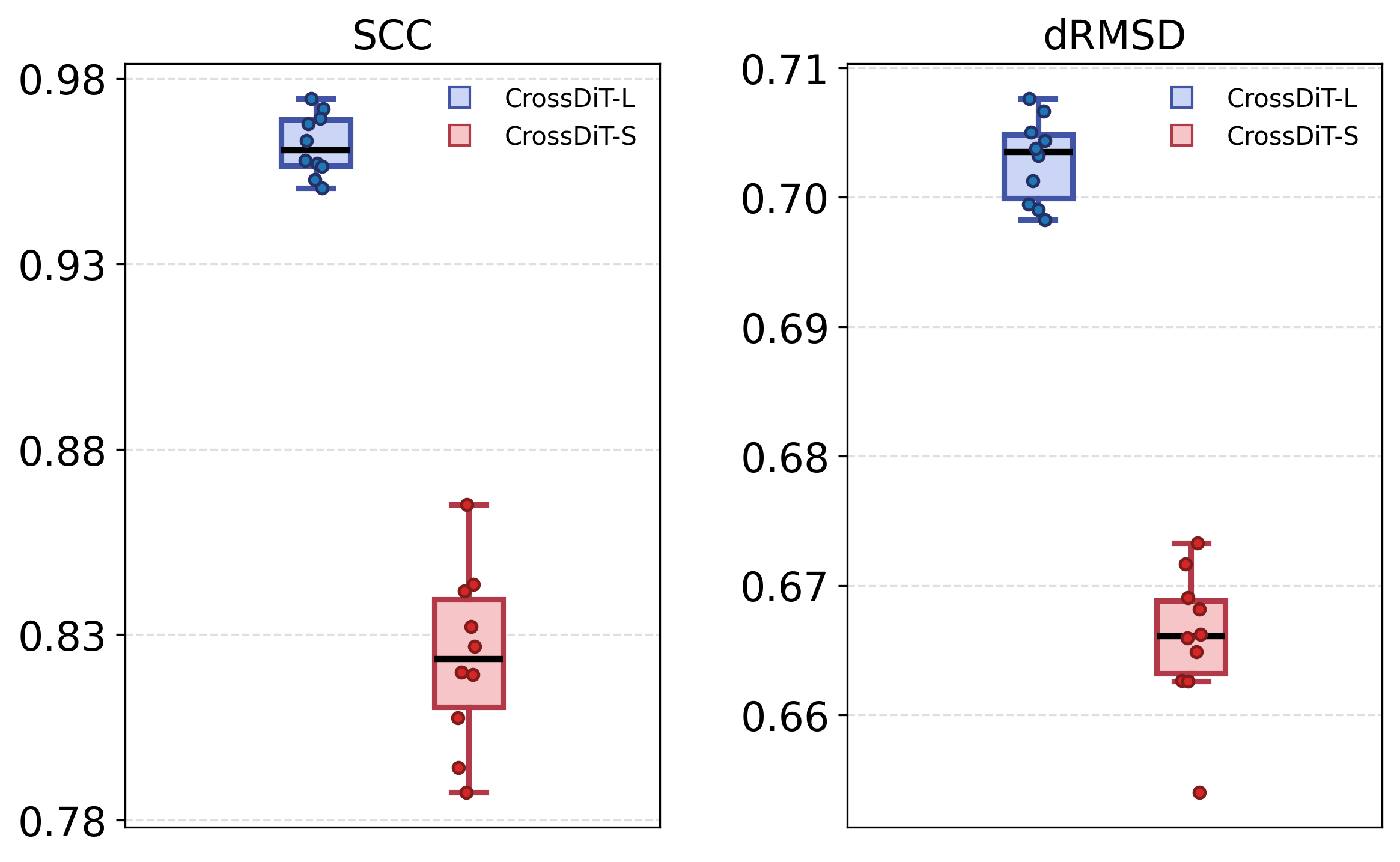} 
        \end{minipage}
        \label{fig:box}
    }
    
  \caption{(a) The P(s) curves of an input Hi-C map from the test set and the corresponding predicted Hi-C maps; (b) SCC and dRMSD computed for each ensemble using all samples within an ensemble.}
  \label{fig:results}
\end{figure}

To assess whether our generated structures reproduce the global polymer-like scaling of contacts with genomic separation, we compared P(s) curves between the input Hi-C map and the contact map calculated by the generated ensemble. Here, P(s) is the average contact frequency between loci separated by genomic distance $s$ (for \textit{E. coli}, using the circular distance $d(i,j) = min(\vert i -j \vert, N-\vert i - j \vert)$). Because P(s) marginalizes over all pairs at the same separation, it provides an intuitive, noise-robust summary of multi-scale compaction and long-range organization that is largely independent of individual pixels. As shown in Fig. \ref{fig:ps}, the predicted P(s) closely matches the input curve across separations, indicating that our method preserves the overall distance-dependent contact regime implied by the guidance Hi-C.

While P(s) validates agreement in the 1D distance–decay profile, it does not fully capture 2D contact-map structure at fixed genomic distances. We therefore additionally reported HiCRep’s stratum-adjusted correlation coefficient (SCC) \cite{yang2017hicrep}, a distance-aware concordance metric that (i) smooths each map to reduce sampling noise, (ii) stratifies contacts by genomic separation (matrix diagonals), (iii) computes correlations within each stratum, and (iv) aggregates them with variance- and sample-size–aware weighting into a single score. SCC complements P(s) by evaluating whether the pattern of contacts within each distance band matches, not only the overall decay. Importantly, SCC is generally more appropriate than a naive Pearson correlation coefficient (PCC) on the full matrix, because PCC is heavily dominated by the strong distance-decay trend and coverage-related effects, whereas SCC explicitly controls for distance by comparing like-with-like strata. Across the 10 test cases, CrossDiT-L consistently achieves high SCC scores (mean 0.962, standard deviation 0.008, range 0.951–0.975). Despite its substantially smaller parameter count, CrossDiT-S also attains competitive performance (mean 0.824, standard deviation 0.022, range 0.787–0.865). These results demonstrate that the proposed method reliably generates ensembles whose implied Hi-C maps closely match the guiding signal beyond the global P(s) decay (as shown in Fig. \ref{fig:box}).

In addition to the similarity, we also evaluated ensemble diversity using the mean pairwise dRMSD (distance-RMSD) computed from the generated 3D coordinates as shown in Fig. \ref{fig:box}. dRMSD measures how much two conformations differ in their internal geometry by comparing all pairwise Euclidean distances within each structure: for a structure $i$, define $D_i(a,b) = \|x_a-x_b\|_2$, and for two samples $i,j$ compute $dRMSD(i,j) = \sqrt{\frac{2}{L(L-2)}\sum_{a<b}(D_i(a,b) - D_j(a,b))^2}$. Averaging dRMSD$(i,j)$ over all pairs in the generated set yields a single diversity score for each condition. This metric is well suited to 3D genome generation because it is invariant to global rigid motions and directly reflects differences in the overall polymer conformation.
To contextualize the magnitude of dRMSD values, we reported a  baseline constructed by taking one generated structure and producing an ensemble via small isotropic Gaussian perturbations of its coordinates; the mean pairwise dRMSD within this perturbed ensemble provides a reference for the variation expected from minor deformations around a single conformation. For the representative condition, the generated ensemble from CrossDiT-L exhibits a mean pairwise dRMSD of 0.700, while CrossDiT-S attains a mean pairwise dRMSD of 0.666; in contrast, the baseline yields a substantially lower value of 0.072. Given a mean bond length of 0.362, the structural diversity of the generated ensembles corresponds to approximately $1.94\times$ and $1.84\times$ the bond length for CrossDiT-L and CrossDiT-S, respectively, whereas the baseline corresponds to only $\sim0.20\times$ the bond length. The results demonstrate that the model does not collapse to a single consensus structure, but instead generates genuinely diverse conformations. 

The results indicate that both generation quality and sample diversity are correlated with model capacity. CrossDiT-L demonstrates that larger models can more effectively learn ensemble-level constraints imposed by Hi-C and support more diverse generation. In contrast, the 45M-parameter CrossDiT-S, while limited in generation quality, is better suited for deployment on more commonly available computing hardware.

\section{DISCUSSION}
These results indicate that our model can effectively follow the guidance of the input Hi-C maps, generating samples that respect the global constraints while maintaining sufficient structural diversity. Compared with traditional approaches that rely mainly on mathematical constraints to infer a single or a small number of representative conformations, our findings support the use of generative models to produce diverse ensembles of fully specified 3D structures at the single-conformation level. This demonstrates that a generative model is a viable and powerful alternative for capturing the variability inherent in chromatin organization under a given Hi-C constraint.

Nevertheless, for the specific biological problem of 3D genome generation, our approach also raises several open questions that merit further discussion.

First, our model architecture is built on CrossDiT, in which conditional information is injected via cross-attention. Cross-attention provides an effective and intuitive mechanism for condition-controlled generation, particularly in our setting where structures must be generated under fixed experimental constraints. In this context, cross-attention can be interpreted as querying sequence representations based on the given condition, thereby retrieving features that are consistent with the imposed Hi-C constraints.

At the same time, many recent multimodal generative models adopt joint-attention mechanisms that allow different modalities to be updated together, with the aim of enhancing representational capacity, learning more expressive interpretations of conditional signals, or improving cross-modal alignment \cite{na2025diffusionadaptivetextembedding, esser2024scalingrectifiedflowtransformers}. This trend motivates further reflection in the context of our task. Specifically, different strategies for conditional injection such as cross-attention, which aligns naturally with fixed physical constraints, versus MMDiT-style joint attention, which may yield richer conditional representations but also risks semantic drift or entanglement \cite{varanka2025zeroshotvideoderainingvideo, toker-etal-2025-padding}, represent an important direction for future investigation. Exploring how these design choices affect interpretability, reliability, and biological plausibility will be a key focus of our subsequent work.

In addition, the current per-bin alignment setting leads to relatively long sequences, resulting in substantial computational and memory overhead. This issue becomes particularly pronounced when introducing MMDiT-style architecture, as joint attention causes the attention map size to grow quadratically with the concatenated sequence length, further amplifying the cost of long sequences. From the perspective of resource efficiency, backbones designed to better accommodate long sequences, such as EDiT \cite{Becker_2025_ICCV}, may therefore represent more suitable choices for chromatin three-dimensional structure generation in future work.

Because symmetric Hi-C matrices introduce redundant computations and the current VAE operates on relatively simple inputs and outputs, more efficient model designs tailored to the data structure are desirable. For example, Hi-C encoders or GNN-based VAEs could reduce redundant processing of symmetric matrices, better capture relational information, and more naturally accommodate more realistic and complex 3D chromatin structures, such as nested replication branches that may arise during DNA replication.

We also aim to extend the framework to support variable-length inputs, enabling application across multiple species. Ultimately, we plan to package the entire system as an open-source tool, with the hope of providing inspiration and a foundation for further research in this area.

\section{CONCLUSION}
In this study, we formulate 3D genome generation from Hi-C data as a conditional generative problem focused on sampling ensembles of conformations rather than producing a single structure. Using molecular dynamics simulations, we construct a dataset of \textit{E. coli} chromosome structure ensembles and their corresponding Hi-C maps, and develop a CrossDiT diffusion transformer with a Hi-C encoder that injects conditional information via cross-attention. The trained model generates diverse three-dimensional structures whose ensemble-averaged contacts match the input Hi-C maps, indicating that diffusion-based generative modeling provides a scalable alternative to deterministic reconstruction methods for capturing structural heterogeneity in genome organization.

\subsubsection*{Acknowledgments}
This work was supported by JST SPRING, Grant Number JPMJSP2108.

\bibliography{gen2_iclr2026_workshop}
\bibliographystyle{gen2_iclr2026_workshop}


\end{document}